%% file: main.tex
\documentclass[letterpaper, 10pt, conference]{ieeeconf}  
\pdfoutput=1

\IEEEoverridecommandlockouts                              

\usepackage{amsmath,bm,times}
\usepackage{graphicx}
\usepackage{cite}
\usepackage{url}

\newcommand{\OUT}[1]{}
\newcommand{\modelname}{Hierarchical Bipartite Action-Transition Network}
\newcommand{\modelacro}{HBATN}

\newcommand{\subtaskacro}{AcTNet}
\newcommand{\subtaskacroplural}{AcTNets}
\newcommand{\componentname}{Mediation Module}

\title{\LARGE \bf
Evaluating Multimodal Interaction of Robots Assisting Older Adults}

\author{Afagh Mehri Shervedani$^{1}$, Ki-Hwan Oh$^{1}$, Bahareh Abbasi$^{2}$, Natawut Monaikul$^{3}$, Zhanibek Rysbek$^{1}$,\\ Barbara Di Eugenio$^{3}$, and Milo\v s \v Zefran$^{1}$
\thanks{$^{1}$A.M. Shervedani, Z. Rysbek, K.H. Oh, and Milo\v s \v Zefran are with the Robotics Lab, Electrical and Computer Engineering Department, University of Illinois Chicago, Chicago, IL 60607 USA.}%
\thanks{$^{2}$B. Abbasi is with the Computer Science Department, California State University Channel Islands, Camarillo, CA 93012 USA.}%
\thanks{$^{3}$N. Monaikul and B. Di Eugenio are with the Natural Language Processing Lab, Computer Science Department, University of Illinois Chicago, Chicago, IL 60607 USA.}%
\thanks{ This work has been supported by the National Science Foundation grants IIS-1705058 and CMMI-1762924.}%
}

\begin{document}

\maketitle

\begin{abstract}
We outline our work on evaluating robots that assist older adults by engaging with them through multiple modalities that include physical interaction. Our thesis is that to increase the effectiveness of assistive robots: 1) robots need to understand and effect multimodal actions, 2) robots should not only react to the human, they need to take the initiative and lead the task when it is necessary. We start by briefly introducing our proposed framework for multimodal interaction and then describe two different experiments with the actual robots. In the first experiment, a Baxter robot helps a human find and locate an object using the Multimodal Interaction Manager (MIM) framework. In the second experiment, a NAO robot is used in the same task, however, the roles of the robot and the human are reversed. We discuss the evaluation methods that were used in these experiments, including different metrics employed to characterize the performance of the robot in each case. We conclude by providing our perspective on the challenges and opportunities for the evaluation of assistive robots for older adults in realistic settings.

\end{abstract}

\input{Introduction.tex}

\input{Literature_Review_MZ.tex}
\input{Framework.tex}
\input{Experiments}
\input{Evaluations}
\input{Discussion}
\input{Conclusion.tex}

\bibliographystyle{IEEEtran}
\bibliography{references,hri_evaluation}

\end{document}

%% file: Introduction.tex
\section{Introduction}
\OUT{KH: Tried to tune the intro and make it focus on the evaluations instead}
With the world population rapidly aging, assistive robots promise to ease the societal burden of care for older adults. The primary focus of care for older adults is on the Activities of Daily Living (ADLs) so that they can continue to live independently, but companionship and socio-emotional support are also important. Increasingly it has been also recognized that helping caregivers may be as important as helping older adults directly.

Evaluation is a critical step in the deployment of assistive robots. Several types of evaluations are typically needed (see also~\cite{tsui_performance_2009,jung_evaluation_2021}):
\begin{itemize}
\item Functionality: does the technology work as intended? For example, does a human action recognition module reach a certain F score?
\item Usability: is the user experience while interacting with the technology satisfactory? For example, can the user interact with the robot using unrestricted instructions, or are they limited to a set of keywords?
\item Effectiveness: does the technology achieve the stated goal? For example, do older people using an assistive robot manage to stay healthier than those that don't?
\end{itemize}
These different types of evaluation increase in complexity, with usability assessment requiring more complex studies than functionality assessment, and effectiveness assessment being significantly more demanding than usability assessment. This is especially true for applications of assistive robots in elderly care, and healthcare in general, where there are many challenges with recruiting subjects, the ability of technology to work in real-life settings, and the length of time needed to assess the health outcomes.


We focus this paper on our experiences with the evaluation of a multimodal interaction manager developed for assistive robots for older adults.
The interactions during various activities of daily living (ADLs) between the human and the robot are expected to be inherently multimodal, such as force exchanges, pointing gestures, haptic-ostensive (H-O) actions, and speech. This is also confirmed by our previously collected corpus of human-human interactions between elderly individuals (elder role: ELD) and nursing students (helper role: HEL) assisting in ADLs~\cite{chen2015RolesofHOAct}. Motivated by this, we proposed a Multimodal Interaction Manager (MIM)~\cite{abbasi2019multimodal} that allows an assistive robot to process the actions of the human, generate appropriate responses, and make progress toward completing the task.  The MIM is described in detail in Sec.~\ref{sec:Framework}. The overview of the implementation and results of our experiments are provided in Sec.~\ref{Experiments}.

During the interaction, the robot may not be able to correctly translate speech to language, and determine other human actions from the readings of the sensors. For instance, the utterance ''cup'' might be misunderstood as ''cop'' which is not an object in the robot's data and results in the robot not being able to complete the task. Furthermore, the gestures of humans, for example pointing to the location, might not be recognized or the pointing direction may not be determined correctly.

Even though the robot may correctly interpret human actions, the robot may fail to correctly respond. For example, the robot may not have a complete representation of the task and could fail to determine what an appropriate response is to a particular human action. In such cases, the robot has to ask the participant to repeat the action until it finds a match in the model it uses for planning. The longer this interaction becomes, the less the user will expect from an assistive robot. We called these as \textit{non-understandings} based on the definition in~\cite{hirst1994repairing}, and thus measuring the rate of unsuccessful attempts is important for evaluating the robot. More details about the evaluations and the discussion are in Sec.~\ref{Evaluations} and Sec.~\ref{Discussion}, respectively.

%% file: Literature_Review_MZ.tex
\section{Related Work}
\label{Related Works}
As assistive robots get more popular and society tends to utilize them more,
the ground metrics become more critical for evaluation. Considering various aspects and applications of different SARs, different evaluation methods should be employed too.

In~\cite{8542489,10.1145/1753846.1754132,8534827}, it is shown that the user’s experience could be refined by adding non-linguistic modalities to robots.
In~\cite{8542489}, the implementation is evaluated on the basis of human participants' answers to the questionnaire covering different metrics.

In~\cite{10.1145/1753846.1754132}, particular metrics are ruled out based on the human participant's words and reactions; and the video recordings of the interaction are analyzed. They also provide the participants with a questionnaire and evaluate the interactions they had with the robot based on their responses to the questionnaire. In~\cite{8534827}, the authors evaluate their approach by implementing it on a robot and measuring the length of the interaction. 

The framework proposed in~\cite{8542489} is evaluated by theoretically analyzing the underlying model before implementing it on a robot. In~\cite{erol1994htn, russell2016artificial}, Hierarchical Task Networks (HTNs) are introduced and evaluated theoretically. 

In~\cite{nooraei2014real}, a middleware system, DiscoRT, is developed and implemented to improve the performance of virtual and robotic conversational agents. The system is evaluated by running experiments on \OUT{one}each virtual and \OUT{one }robot agent and investigating the conceptual aspects of the experiments. \OUT{In order}To analyze the performance of their Interactive Hierarchical Task Learning Algorithms, the authors in~\cite{mohseni2015interactive} run simulations \OUT{experiments by having}where human subjects interact with the simulated robot environment through their graphical user interface. They extract and measure some objectives associated with the task for evaluation\OUT{ purposes}. Similarly, in~\cite{gavsic2015distributed} the authors also run \OUT{stimulative experiments}simulations with specific metrics for evaluating\OUT{ procedure on} their Hierarchical Distributed Dialogue Architecture.

In~\cite{peng2017composite}, the proposed Hierarchical Deep Reinforcement Learning framework is evaluated by reporting the success rate, the average number of turns between the user and the agent, and the reward from the simulation experiments. In~\cite{mcmillen2006distributed}, the Distributed Play-based Role Assignment Algorithm is developed and tested by implementing it on a distributed team of robots for the RoboCup four-legged league. The task-completion time is used as an evaluation metric.

In the literature, however, we observe a lack of separate methodologies for evaluating the theoretical framework and implementation. In this work, we propose adopting different evaluation methods for different aspects of ASRs.

%% file: Framework.tex
\section{Multimodal Interaction Manager Framework}
\label{sec:Framework}

\begin{figure}[t]
\centering
\includegraphics[width=0.75\columnwidth]{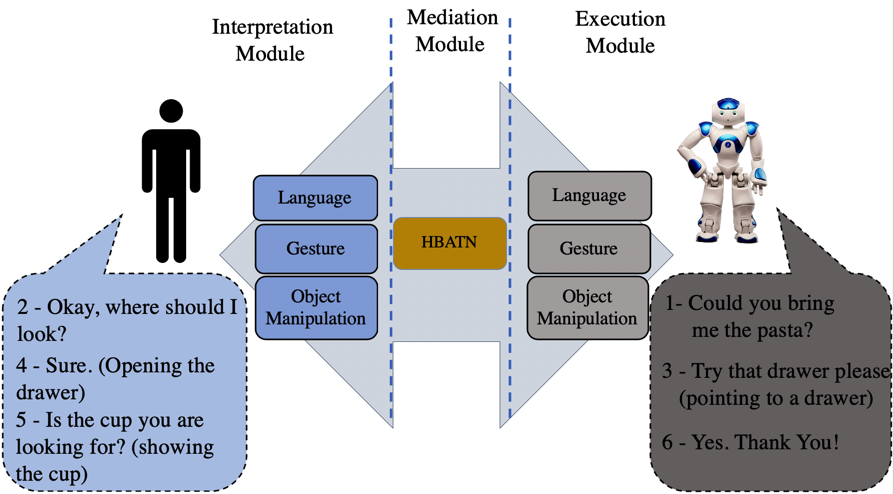}
\caption{The architecture for multimodal human-robot interaction. The figure is taken from\cite{monaikul2020role}.}
\label{fig:HRI}
\vspace{-5mm}
\end{figure}

As shown in Fig~\ref{fig:HRI}, the Multimodal Interaction Manager (MIM) consists of three components: (a) the interpretation module, which interprets multimodal actions of the human observed by the robot; (b) the mediation module, which determines the action of the robot in response to the human; and (c) the execution module, which executes the action of the robot. The task that was studied in detail in our work was the \textit{Find} task, an interaction scenario in which a human and a robot work together to find an object in the environment. The core of our framework is Hierarchical Bipartite Action-Transition Networks (HBATNs) that model both agents simultaneously to maintain the state of a task-driven multimodal interaction and plan subsequent robot moves.

\OUT{In this section, we give a brief overview of our previous framework and the corpus on which our \modelacro{} was developed.}

The ELDERLY-AT-HOME corpus~\cite{chen2015roles}, a publicly available corpus of human-human multimodal interactions, involves performing assisted ADLs, such as putting on shoes and preparing dinner. We developed the framework of \modelacro{} on a subcorpus consisting of the interactions related to the \emph{Find} task. That is, the elderly participant (ELD) would ask for an object, and the helper (HEL) would try to find it by asking follow-up questions.

\OUT{We formulated}The \emph{Find} task can be decomposed into \OUT{as} a set of subtasks \OUT{with the goal of}to identify two main unknowns: the target object ($O$) and its location ($L$)\OUT{ of the object}. \OUT{As a result, we modeled }The four main subtasks are determining the desired object type (\textit{Det($O_{T}$)}), determining a potential location to check (\textit{Det($L$)}), opening the location (\textit{Open($L$)}), and determining the actual object (\textit{Det($O$)}). These are modeled as Action-Transition Networks (\subtaskacroplural{}).

The \subtaskacro{} is a bipartite graph representing the states of both participants \OUT{(ELD and HEL)}and their possible multimodal actions, which are defined as vectors consisting of linguistic features (the \emph{dialogue act} (DA)~\cite{chen2015roles} of the utterance and object or location words) and physical features (pointing gestures or \emph{haptic-ostensive} (H-O) actions). The \modelacro{} encompass\OUT{ing} these \subtaskacroplural{} \OUT{then }allowing a robot to not only infer the state of its partner but also to plan its next action accordingly.

We generalized our model\OUT{ even more} to enable the robot to be either the ELD or HEL by decomposing \OUT{each subtask into finer subtasks, which we call \emph{primitive subtasks}}the subtasks into what we call \emph{primitive subtasks}. In this new formulation, \emph{Det($O_T$)} and \emph{Det($L$)} \OUT{consist of }establishes the object type \OUT{or}and its location (\emph{Estab}), potentially followed up by verification (\emph{Verify}) or \OUT{follow-up}questions specifying for more information (\emph{Spec}), and\OUT{ that} \emph{Det($O$)} \OUT{consists of }confirms the presence or absence of the desired object (\emph{Finish}) in the current location or verify a physical object with the partner.

Subsequently, a classifier was developed to allow the robot to determine what is the primitive subtask that the interaction is currently in. The proposed classifier \OUT{has the potential for}automatically annotates multimodal interaction data for our primitive subtasks\OUT{, which}. This can \OUT{ then} be used in turn to learn the topologies of each subtask by extracting \OUT{consecutive } sequences of moves belonging to the same subtask and using well-established techniques for learning\OUT{, for example,} like Markov models. It is \OUT{can be used in}implemented on the MIM to \OUT{help }infer the state of the human partner by comparing the observed human action with all possible actions in the \modelacro{} with preference given to those in the predicted subtask. A comprehensive demonstration of the interaction \OUT{while completing the \emph{primitive subtasks}} can be found in~\cite{monaikul2020role}.
\OUT{KH: I think this might be too much}
\OUT{By further abstracting the model into \emph{primitive subtasks}, more opportunities for switching roles or initiative arise throughout the interaction, and these primitive structures can be more easily incorporated as building blocks of other tasks.} 


%% file: Experiments.tex
\section{Implementation and Experiments}
\label{Experiments}
To test the full MIM depicted in Fig.~\ref{fig:HRI}, we implemented several components in both experiments to recognize and understand multimodal human actions (the Interpretation Module), and to generate robot actions (the Execution Module).
In this section, we focus on the results and the evaluations.

\begin{table*}[t]
\vspace{2mm}
\centering
\begin{tabular}{||cccccccccc||}

 \hline
Avg. & Avg. \# & Successful & Non- &  &  & Wrong & SSRE \& Wrong & Wrong & Avg. User \\
Duration & Turns & Trials & Understandings & WER & SSREs & Pointing & Pointing & DAs & Rating\\
 \hline\hline

1m 45s & 15.6 & 85.7\% & 11.7\% & 16.3\% & 23.4\% &  28.9\% & 1.2\% & 11.1\% & 4\\[0.5ex]

\hline
\end{tabular}

\caption{Performance results of the MIM on the \textit{Find} task with Baxter as the HEL. The table is taken from\cite{abbasi2019multimodal}.}
\label{table:usersample}
\vspace{-3mm}
\end{table*}

\subsection{Robot as the HEL}
To evaluate the performance of our initial Multimodal Interaction Manager, the framework was implemented on Baxter Robot from Rethink Robotics. In this experiment, Baxter participated as HEL and human participants acted as ELD in the \textit{Find} task. 
The human participant would give instructions to Baxter \OUT{in order }to guide it through the find task \OUT{On the other hand}while Baxter robot would ask questions regarding the object and its location.\OUT{ in which the object could be.}

In this experiment, \OUT{the human acting as }the ELD (human) rarely performs H-O actions. As a result, \OUT{the focus of }the Interpretation Module \OUT{is}focuses only on interpreting human speech and \OUT{pointing }gestures; and this is composed of three primary parts: a speech-to-text component, a pointing gesture recognition component, and the \textit{Dialogue Processing \& Modality Fusion} component for processing the utterance and gesture together.

\OUT{Baxter acting as }The HEL (Baxter) \OUT{is supposed to and can}should perform pointing actions as well as H-O actions. Baxter \OUT{is also able to}can also communicate to the ELD through generated speech. Therefore, the Execution module in this experiment performs H-O actions, pointing actions, and speech; and is composed of multiple subcomponents: pointing, H-O action, speech generation, and object recognition component that enables the robot to determine \OUT{what object is located in a specific location}the object and its location. The output of the \componentname{} is an action vector defining the robot's next move; each execution component uses this vector to perform its respective action.

\subsection{Robot as the ELD}
\OUT{Furthermore, }To evaluate the feasibility of switching roles, we \OUT{again }implemented the MIM with the refined \modelacro{}. In this experiment, the NAO robot acted as the ELD and human participants acted as HEL in the interaction. Utilizing a different robot in this experiment confirms the fact that our framework is platform-independent. 

In this experiment, \OUT{the human acts as}the HEL (human) performs H-O and pointing actions as well as communications through speech. As a result, our Interpretation Module has \OUT{to be able }to interpret not only the human's speech and pointing gestures but also their H-O actions. The Interpretation Module is composed of the followings\OUT{ subcomponents}: pointing gesture recognition, H-O action recognition, a speech-to-text component, an object recognition component, and most importantly the \textit{Dialogue Processing \& Modality Fusion} component. The last component performs DA and subtask classification, combines the results \OUT{to create}creating an input action vector, and transfers it to the Mediation Module.

\OUT{NAO acting as }The ELD (NAO) only needs to perform pointing gestures and speech. Thus, the Execution Module only takes care of\OUT{ only} the pointing gesture and speech, \OUT{being}and thus it contains\OUT{ composed of } two components: pointing gesture execution and speech generation.

%% file: Evaluations.tex
\section{User Study and Evaluations}
\label{Evaluations}

\subsection{Experiment on Baxter}
In a preliminary user study, seven participants were recruited to interact with Baxter. Each subject performed 4 trials to find one of the four objects, giving a total of 28 trials of \textit{Find} task. 
Baxter would help the participants locate the object by talking and pointing. No script was provided to the participants.

\begin{table*}[t]
\vspace{2mm}
\centering
\begin{tabular}{|cccccccc|}

 \hline
 Avg. \# & Successful & Non- &  STT& DA & H-O & Pointing  & Subtask\\
Moves  & Trials & Understandings & Accuracy & Accuracy & Accuracy &  Accuracy& Accuracy\\
 \hline

19 & 84.8\% & 32.6\% & 84.8\% & 57\% &  83.1\% & 96\% & 49.3\% \\[0.5ex]

\hline
\end{tabular}

\caption{Performance results of the MIM on the \textit{Find} task with NAO as the ELD. The table is taken from\cite{monaikul2020role}.}
\label{table:usersample1}
\vspace{-4mm}
\end{table*}

\begin{table*}[t]
\vspace{2mm}
\centering
\begin{tabular}{|ccccccc|}

 \hline
 DA &  Speech  & H-O&  Pointing & Subtask  & DA \& Subtask & Model\\
 Failure & Failure & Failure & Failure & Failure & Failure & Failure\\

\hline
 55.5\% & 43.4\% & 22.2\% &  2\% & 92.9\% & 51.5\% & 14.1\% \\

\hline
\end{tabular}

\caption{Percentage of non-understandings in which errors in each component occur. The table is taken from\cite{monaikul2020role}.}
\label{table:usersample2}
\vspace{-4mm}
\end{table*}

The evaluation has been done by reporting: 1) Average length of the interactions as the mean duration and the mean number of moves; 2) The percentage of successful trials, (trials in which Baxter continued the interaction when the object was already found are counted as failed trials); 3) \textit{Non-understanding} \OUT{consistent with the definition in \cite{hirst1994repairing}, mentioned in the Intro}percentage of turns (throughout all trials) where the Baxter's interpretation of the human action can not be \OUT{matched to an action}found in \modelacro{} and Baxter needs to ask the participant to repeat their action; 4) The word error rate (WER)~\cite{jurafsky2008speech} and serious speech recognition errors (SSREs)~\cite{abbasi2019multimodal} for checking the speech recognition component of the Interpretation Module; 5) The percentage of wrong pointing gestures that were either not recognized or not tagged with the correct intended location for gesture recognition component; 6) The percentage of wrong classified DAs compared with the manually-labeled DA tags; 7) Overall quality of the interaction by asking the participants to rate their experience on a 5-point Likert scale~\cite{abbasi2019multimodal}.

\subsection{Experiment on NAO}
In another preliminary user study, six participants were recruited to interact with NAO. Each subject performed 5 to 6 trials with a total of 28 \textit{Find} task trials. Participants would help NAO find the object it had in mind from a specific location.

Similar to the previous experiment, various metrics are reported for the evaluation: 1) The percentage of successful trials; 2) \emph{Non-understanding} percentage of human turns (throughout all trials) in two categories where \OUT{either }NAO asks the participant to repeat their action, and NAO fails to answer the participant's question or to follow their instruction; 3) The percentage of non-understandings in which various components make mistakes, particularly if an action is not accounted in the \modelacro{}; 4) The speech-to-text (STT) accuracy~\cite{monaikul2020role}; 5) The accuracy of DA classifier; 6) The accuracy of H-O action recognition; 6) The Accuracy of pointing gesture recognition; 7) The accuracy of subtask classifier.

%% file: Discussion.tex
\section{Discussion}
\label{Discussion}
For a detailed discussion of the results of the experiments, the reader is referred to~\cite{abbasi2019multimodal,monaikul2020role}. The focus of this paper is on how our evaluation methods could practically be utilized in evaluating assistive robots.

One vital aspect of an interaction between a human and a robot is the duration of the interaction (a type of usability). As we pointed out before, one important application for assistive robots is to help older adults. The longer one interaction takes the more frustrated the human becomes. It also could highly affect the efficiency of task completion because if the older adult gets tired they are less interested in engaging in the task. \OUT{We argue}Based on our experiment, we can declare that the average duration of the interactions \OUT{ of our user studies }is less than the frustration threshold of humans. Moreover, the average user Likert ratings of 4 out of 5 \OUT{reported in our results }greatly supports our argument\OUT{ in this regard}.
However, most of the subjects were young people, and the ratings may decrease when more elderly subjects are involved.
\OUT{However, }Since our studies remain at a more theoretical level and are not immediately relevant for applications in the real world, evaluation with older adults remains part of our future work.


An alternative aspect of usability is the overall success rate of the interactions between humans and robots. 
Humans tend to expect an assistive robot to act similarly to a human.
Since humans are extremely adept at completing interactions, they expect similar performance from a robot. The success rates reported in Tables~\ref{table:usersample} and~\ref{table:usersample1} show that our proposed framework achieves very good performance in this regard too.

Evaluating the components of \OUT{one}a system \OUT{ is consisted of} can reveal many hidden issues.
Each sub-system is closely connected to its adjacent components and it is likely that an error made by one component dramatically affects the overall performance of the robot and the success of the interaction. Evaluating components thus provides important insight for the robotic community by identifying possible failure points and establishing the relative importance of different components. For instance, the detailed results of sub-component accuracy/ failure we reported in Tables~\ref{table:usersample}, ~\ref{table:usersample1}, and~\ref{table:usersample2} explain unsuccessful trials and non-understanding turns~\cite{abbasi2019multimodal},~\cite{monaikul2020role}. In particular, DA classifier accuracy of 57${\%}$ contributes to a non-understanding rate of 32.6${\%}$ in the NAO experiment.
\OUT{KH: Perhaps pointing out Table 3 and showing that it helps analyze where most of the errors are coming from might be more relevant?}

The discussion above raises the question of which is more important: the theoretical framework itself or the implementation? Will a superior implementation of a mediocre framework outperform a mediocre implementation of a superior framework? Clearly, the theoretical framework needs to be implemented on a real robot, or at least in a simulation, to be evaluated. However, the limits on time and resources frequently prevent researchers from spending sufficient effort on implementation.

\OUT{KH: Changed the argument based on my opinion. However, feel free to modify it if you want to add yours too.}
\OUT{We argue that although to evaluate the framework itself implementing the setup is essential, the evaluation of the framework should be separated from the implementation. Researchers in the community have different areas of interest. Also, limitations in time and resources do not allow every researcher to present the best implementation for their studies. Thus, theoretical contributions which tend to push the boundaries of the state-of-the-art frameworks should be evaluated separately from application-focused studied which try to propose the best implementations.}

Finally, while evaluation in real-world applications is clearly the ultimate test, it is becoming increasingly common to evaluate assistive robots in simulations. How to properly interpret the simulation results and reduce the cost of real-world evaluation remains an important topic for future research.



%% file: Conclusion.tex
\section{Conclusion}
\label{conclusion}
In this work, we summarized our previous studies of assistive robots capable of multimodal interaction and described in detail the metrics used for their evaluation. These metrics should be of general interest and we hope that our insights can benefit other researchers in the area of assistive robots. We provided the motivation for using various metrics and showed that defining metrics tailored to evaluation of different components of the overall system can help explain its overall performance. 